%% file: acl_latex.tex
\pdfoutput=1

\documentclass[11pt]{article}
\usepackage{float}
\usepackage[final]{acl}

\usepackage{times}
\usepackage{latexsym}
\usepackage{tcolorbox}
\usepackage[T1]{fontenc}
\usepackage{booktabs}
\usepackage{xcolor}
\usepackage{algorithm}
\usepackage{algpseudocode}

\usepackage{amsmath}

\usepackage[utf8]{inputenc}
\usepackage{environ} 
 
\NewEnviron{NORMAL}{% 
    \scalebox{0.95}{$\BODY$} 
} 

\usepackage{microtype}
\usepackage{enumitem}

\usepackage{inconsolata}

\usepackage{graphicx}

\usepackage{multirow}

\usepackage{bbding}
\usepackage{pifont}
\usepackage{wasysym}
\usepackage{amssymb}

\usepackage{array}

\title{ \textit{SWITCH}: Studying with Teacher for Knowledge Distillation of Large Language Models}
\author{
Jahyun Koo\textsuperscript{1} \hspace{1.0cm} 
Yerin Hwang\textsuperscript{1}\hspace{1.0cm} 
Yongil Kim\textsuperscript{2} 
\\ 
\hspace{0.5cm} 
\textbf{Taegwan Kang}\textsuperscript{2}  \hspace{0.4cm} 
{\bf Hyunkyung Bae\textsuperscript{2}} \hspace{0.4cm}  
{\bf Kyomin Jung\textsuperscript{1,3,4$\dagger$}} \\
  $^{1}$IPAI, Seoul National University \hspace{1.0cm} 
  $^{2}$LG AI Research\\
  $^{3}$Dept. of ECE, Seoul National University\hspace{0.7cm} 
  $^{4}$SNU-LG AI Research Center\\
  \texttt{\{koojahyun, dpfls589, kjung\}@snu.ac.kr}\\
  \texttt{\{yong-il.kim, taegwan93.kang, hkbae\}@lgresearch.ai}
  }
\begin{document}
\maketitle
\begin{abstract}
Despite the success of Large Language Models (LLMs), they still face challenges related to high inference costs and memory requirements. To address these issues, Knowledge Distillation (KD) has emerged as a popular method for model compression, with the use of student-generated outputs (SGOs) as training data being particularly notable for reducing the mismatch between training and inference. However, SGOs often produce noisy and biased sequences, which can lead to misguidance from the teacher model, especially in long sequences. To mitigate these challenges, we propose \textbf{SWITCH} (\textbf{S}tudying \textbf{WI}th \textbf{T}ea\textbf{CH}er for Knowledge Distillation), a novel approach that strategically incorporates the teacher model during the student’s sequence generation. SWITCH identifies discrepancies between the token probabilities of the teacher and student models, allowing the teacher to intervene selectively, particularly in long sequences that are more prone to teacher misguidance. Extensive experimental results across three model families and five instruction-following datasets show that SWITCH surpasses traditional KD methods, particularly excelling in the generation of long sequential data.
\end{abstract}

\section{Introduction}
\input{text/1_introduction}

\section{Methodology}
\input{text/3_methodology}

\section{Experiments}

\input{text/4_experiments}

\section{Analysis}

\input{text/5_analysis}

\section{Related Works}
\input{text/2_relatedworks}

\section{Conclusion}
\input{text/6_conclusion}

\section*{Limitations}
While SWITCH achieves state-of-the-art performance, it faces limitations in terms of computational resources due to the involvement of larger teacher models in the generation process. However, our approach significantly improves performance, and the additional training time does not affect the deployment of the compressed language model. This aligns with the overall goal of knowledge distillation, which is to produce a smaller, more efficient model for deployment.

\section*{Ethics Statement}
Pre-trained language models come with inherent ethical and social risks~\cite{bommasani2021opportunities,weidinger2021ethical, kim-etal-2024-lifetox, koh-etal-2024-llms}. Additionally, model compression techniques can amplify existing biases in the original models~\cite{hooker2020characterising,goncalves-strubell-2023-understanding}. SWITCH employs pre-trained language models and applies compression methods, making it susceptible to these risks. However, since most compression studies use pre-trained models, these risks are general and not specific to SWITCH.
%% add
In this work, we used publicly available datasets, ensuring they were applied in accordance with their original purposes. Our evaluations were conducted using the official API, and all models and source codes were obtained from their respective official repositories.

While writing this paper, we used an AI assistant to help draft and refine sentences at the sentence level, as well as to check for grammatical errors.

\section*{Acknowledgments}
This work was supported by LG AI Research. This work was partly supported by the Institute of Information \& Communications Technology Planning \& Evaluation(IITP)-ITRC(Information Technology Research Center) grant funded by the Korea government(MSIT)(IITP-2025-RS-2024-00437633, 30\%), Institute of Information \& communications Technology Planning \& Evaluation (IITP) grant funded by the Korea government(MSIT) [RS-2021-II211343, Artificial Intelligence Graduate School Program (Seoul National University) \& RS-2021-II212068, Artificial Intelligence Innovation Hub (Artificial Intelligence Institute, Seoul National University)], and the BK21 FOUR program of the Education and Research Program for Future ICT Pioneers, Seoul National University in 2024.
K. Jung is with ASRI, Seoul National University, Korea.
\bibliography{custom}

\clearpage

\appendix

\input{text/7_appendix.tex}

\end{document}

%% file: text/1_introduction.tex
Despite the strong performance of large language models (LLMs,~\citealp{alpaca,openai2023gpt4}), their immense scale incurs high resource demands, leading to efforts to compress them while preserving performance \cite{hsieh2023distilling,jiang2023lion,zhong2024revisiting}.
To address this issue, Knowledge Distillation (KD, \citealp{hinton2015distilling}) has emerged as a promising method for compressing LLMs, aiming to transfer knowledge from a large teacher model to a smaller student model without substantial loss in performance. While traditional KD techniques have focused on natural language understanding \citep{sanh2019distilbert, mirzadeh2020improved}, recent research has shifted towards applying KD to natural language generation~\cite{lin2020autoregressive,wen2023f}.
In this vein, methods utilizing Student-Generated Outputs (SGOs)—where the student model's own outputs are used as training sequences—have been proposed to improve text generation performance. These approaches help mitigate the training-inference mismatch~\cite{bengio2015scheduled}, leading to notable performance improvements \cite{gkd,minillm}.

However, SGO methods primarily emphasize efficient training of the student model while often overlooking the necessity of precise guidance from the teacher model. This approach accounts for the training-inference mismatch from the student's perspective but neglects the analogous mismatch from the teacher’s perspective, potentially resulting in misguidance during training \cite{ko2024distillm}. This misguidance is due to the inherent capacity gap between the teacher and student models, and the problem becomes more pronounced as the gap widens. Furthermore, such misguidance is a significant concern, as the KD method is based on the assumption that the teacher offers reliable guidance. Specifically, it causes the student to receive high loss penalties for correct predictions and low penalties for incorrect ones, resulting in inaccurate knowledge transfer.

\begin{figure}[t]
\centering
\includegraphics[width= 1\columnwidth]{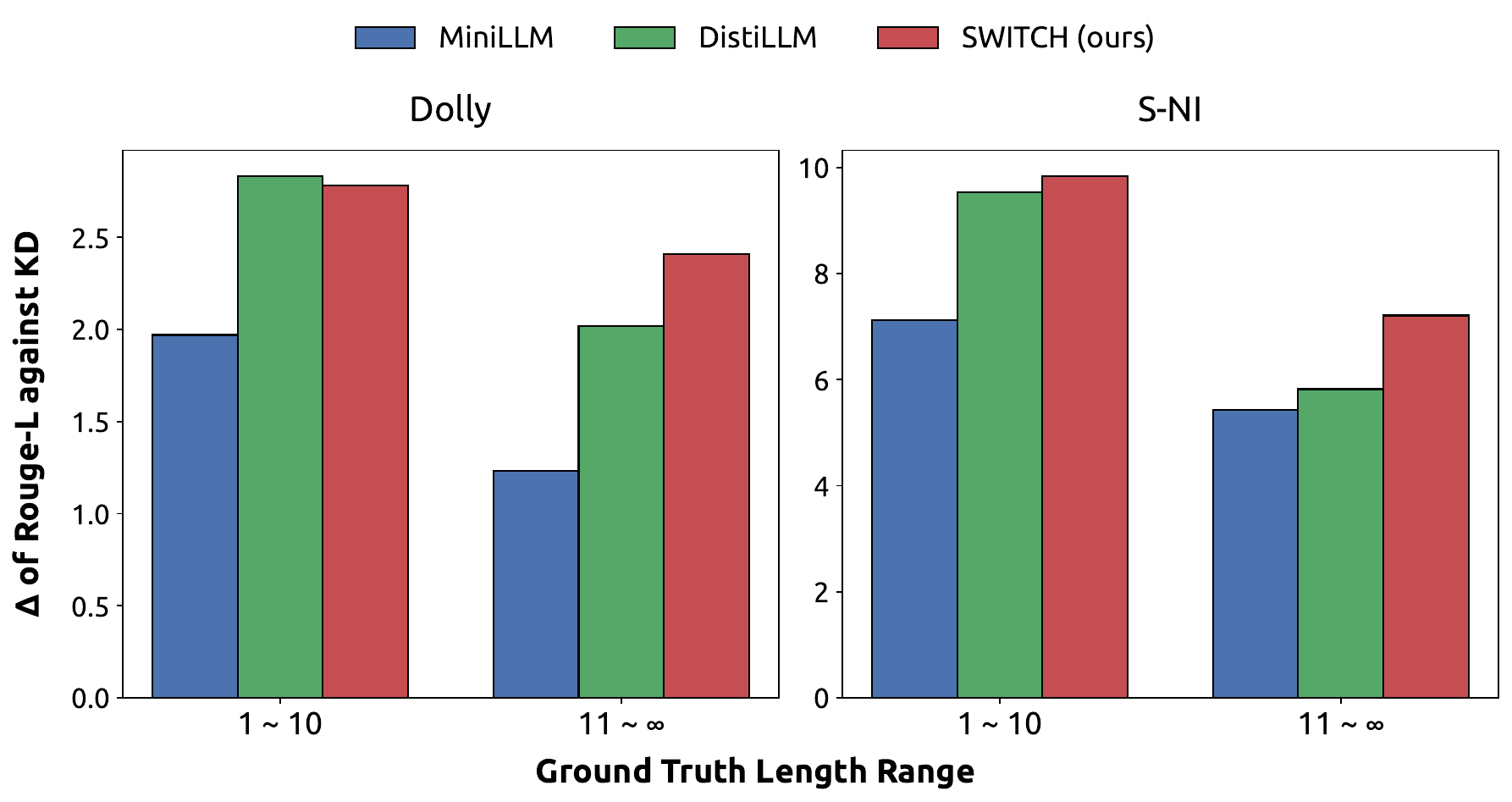} 
\caption{Rouge-L scores of the distilled models against KD. Dataset split by ground truth length. }
\label{figure1}
\vspace{-5mm}
\end{figure}

Moreover, the autoregressive nature of language models exacerbates this issue, as biases from the student model can accumulate over long sequences \cite{arora2022exposure}. This problem manifests in instruction-following tasks, which are a general form of sequence generation \cite{ouyang2022training}. We observe that existing KD methods utilizing SGOs struggle with long sequential data, as shown in Figure~\ref{figure1}, underscoring the need for more sophisticated strategies when applying SGOs in KD frameworks. Therefore, excessive reliance on outputs generated solely by the student model may lead to misguidance from the teacher.

To address this issue, this paper proposes an innovative approach, termed \textbf{SWITCH} (\textbf{S}tudying \textbf{WI}th \textbf{T}ea\textbf{CH}er for knowledge distillation), which strategically incorporates the teacher model into the generation of student sequences to ensure accurate guidance from the teacher. By detecting discrepancies in the probability distributions of the next token between the teacher and the student, our method selectively generates tokens with the teacher model instead of the student model. This teacher intervention is particularly effective for long sequences, where the risk of cumulative errors and the following misguidance becomes more significant. To manage this, SWITCH leverages an exponentially decaying threshold that increases the teacher's involvement as the sequence progresses, preventing misguidance in long outputs.

Through comprehensive experiments, we demonstrate that the SWITCH method outperforms existing baseline approaches. This performance improvement has been verified across five instruction-following benchmarks and three model families, and it remains effective across different student model sizes. Notably, the performance gains achieved by SWITCH increase as the size difference between the student and teacher models grows, suggesting that our method effectively mitigates the misguidance issues caused by excessive reliance on student-generated outputs. Furthermore, we observe significant performance improvements in generating long sequences. This highlights the critical role of our method in applying KD for long sequences.

To summarize, our contributions are three-fold:
\begin{itemize}

\item We propose SWITCH, a novel approach that utilizes selective intervention of the teacher model to mitigate misguidance from student-generated outputs. 

\item We demonstrate that SWITCH achieves state-of-the-art performance across various benchmarks and model sizes.

\item Our method particularly excels when there is a substantial size difference between the student and teacher models and when handling long sequences.

\end{itemize}

%% file: text/3_methodology.tex
\input{src/fig_picture}
In this section, we introduce \textbf{SWITCH}, a novel approach designed to enhance the knowledge distillation process for language models. We begin by formalizing the problem of sequence-level knowledge distillation and then delve into the specifics of SWITCH, explaining how it addresses the limitations of traditional student-generated output (SGO) methods. Our methodology emphasizes the strategic involvement of the teacher model during sequence generation, aiming to reduce the accumulation of errors and improve the overall performance of the student model.

\subsection{Preliminaries}
In knowledge distillation for language models, a smaller student model \( q \) learns to emulate a larger teacher model \( p \). Given a prompt \( x \) and ground-truth sequence \( y = (y_1, y_2, \dots, y_{_T}) \), the training objective for the student model is to minimize the divergence \( D \) of token-level distribution between the student and teacher.

\begin{equation} \label{eq1}
\begin{split}
\mathcal{D}(p \parallel q)(y|x) = \sum_{t=1}^{|y|} \sum_{y_t \in V}\mathcal{D}(p \parallel q)(y_t | {y}_{<t},x)
\end{split}
\end{equation}

where \( y_{<t} \) denotes the sequence of tokens generated up to time \( t-1 \), and $V$ is the vocabulary.

Meanwhile, in SGO methods, the student model generates sequences based on its own probability distribution:
\begin{equation} \label{eq2}
\begin{split}
y_t \sim q(\cdot \mid y_{<t}, x)
\end{split}
\end{equation}

Using student-generated sequence, the student learns by comparing its output distribution to that of the teacher, aiming to minimize the divergence between their respective distributions. The rationale behind the use of SGO is to use its own outputs to reduce training-inference mismatch.

However, since the student model \( q \) typically has less capacity than the teacher model \( p \), it tends to produce noisier and more biased sequences. Due to the autoregressive nature of sequence generation, these errors can accumulate over time, leading to significant divergence from the teacher's behavior.

\subsection{SWITCH}

To overcome the limitations of SGO methods, we propose SWITCH, which strategically incorporates the teacher model into the sequence generation process. The key idea is to selectively switch from the student to the teacher model to generate the next token when significant discrepancies between their distributions are detected in order to minimize distribution mismatch from the teacher model's perspective, thereby reducing misguidance. To address accumulated bias in long sequences, SWITCH increases teacher involvement as the sequence progresses. This selective involvement aims to minimize the misguidance from the teacher model caused by accumulated errors from SGO while preserving the benefits for the student to learn from its own outputs. Figure~\ref{figure_picture} provides an overview.

\paragraph{Measuring Distribution Discrepancy for Selective Token Generation}

To determine when to involve the teacher model, we need a reliable measure of the discrepancy between the student and teacher distributions for the next token. We employ the Jensen-Shannon divergence (JSD), which provides a symmetric and bounded measure of divergence between two probability distributions.

The JSD between the teacher distribution \( p(\cdot \mid x_{<t}) \) and the student distribution \( q(\cdot \mid x_{<t}) \) is defined as:
\begin{equation} \label{eq3}
\begin{split}
\NORMAL{\text{JSD}(p \parallel q) = \frac{1}{2} D_{\text{KL}}(p \parallel m) + \frac{1}{2} D_{\text{KL}}(q \parallel m)}
\end{split}
\end{equation}

where \( m = \frac{1}{2}(p + q) \) is the average distribution, and \( D_{\text{KL}} \) denotes the Kullback-Leibler divergence (KLD). The JSD ranges between 0 and 1, facilitating a consistent thresholding mechanism.

At each time step \( t \), we compute the JSD between the student and teacher distributions conditioned on the current context. If the divergence exceeds a predefined threshold \( \tau_t \), we switch from using the student model to the teacher model for generating the next token:

\begin{equation} \label{eq4}
\begin{split}
y_t = \begin{cases}
\text{Sample from } q(\cdot \mid y_{<t}, x), & \text{if } \text{JSD}_t \leq \tau_t \\
\text{Sample from } p(\cdot \mid y_{<t}, x), & \text{if } \text{JSD}_t > \tau_t
\end{cases}
\end{split}
\end{equation}

This strategy allows the student to produce sequences that are reliable enough for the teacher to interpret and offer accurate guidance, as the teacher only intervenes when there is a significant deviation between the two distributions.

\paragraph{Handling Accumulated Bias with Exponentially Decaying Threshold}

To address the accumulation of bias in long sequences due to the autoregressive nature of the model, we introduce an exponentially decaying threshold for the JSD:
\begin{equation} \label{eq5}
\begin{split}
\tau_t = \tau_0 \cdot e^{-\lambda t}
\end{split}
\end{equation}

where:
\begin{itemize}
    \item \( \tau_0 \) is the initial threshold at \( t = 0 \), set to 1.
    \item \( \lambda \) is the decay rate controlling how quickly the threshold decreases over time.
\end{itemize}

This approach encourages the student model to learn from its own outputs at the beginning of the sequence. As the sequence progresses, the threshold \(\tau_t\) decreases exponentially, increasing the likelihood of teacher involvement in later tokens. Since noises of student generations accumulate in long sequences, a lower threshold in subsequent steps ensures that the teacher model can step in to maintain correct guidance. The exponentially decaying threshold thus balances the need for the student to learn from their own outputs with the necessity of preventing error accumulation in long sequences.
We summarize the SWITCH algorithm in Algorithm~\ref{switch}

\input{src/alg}

%% file: src/fig_picture.tex
\begin{figure*}[t!]
\centering
\includegraphics[width=\textwidth]{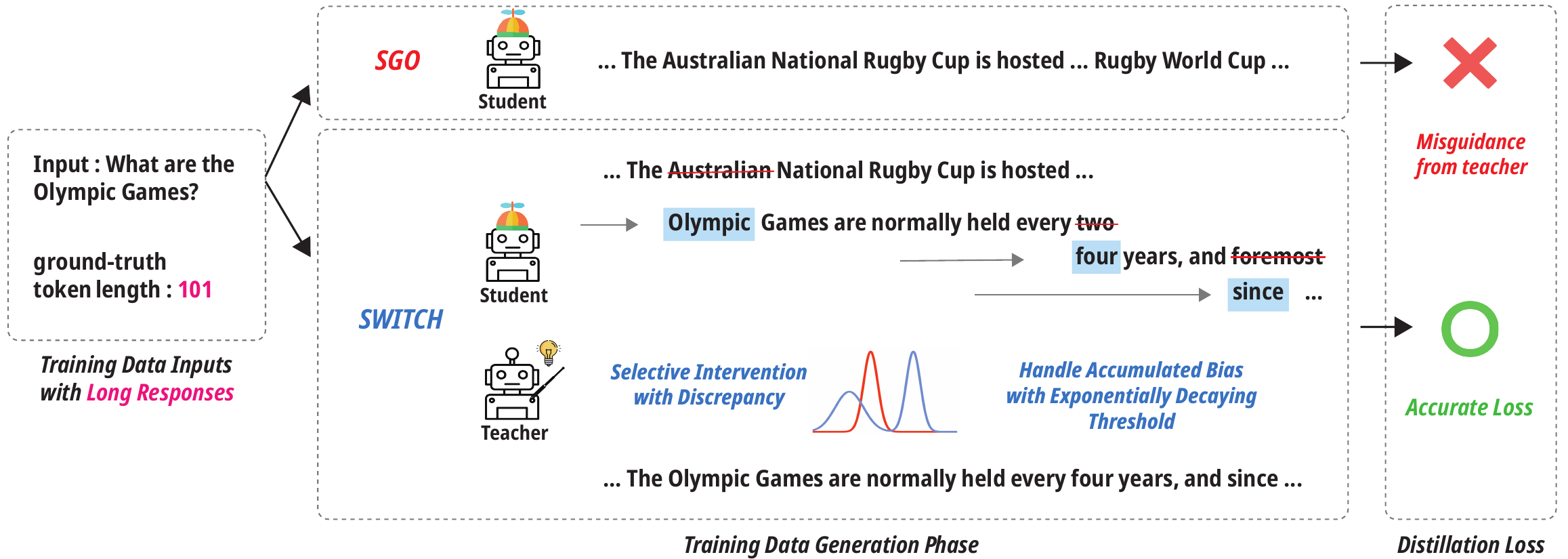} 
\caption{An overview of our SWITCH method. To mitigate misguidance, SWITCH selectively intervenes the generation process of SGO using distribution discrepancy. More intervention is given as the sequence gets longer to balance the benefits of student learning from their own outputs and the mitigation of teacher misguidance. }
\label{figure_picture}
\end{figure*}

%% file: src/alg.tex
\begin{algorithm} 
\small 
\caption{\textsc{SWITCH}} 
\label{switch} 
\begin{algorithmic} 
\Require Dataset $(X,Y)$ of prompts and gt responses 
\State \quad \quad \ teacher model $p$, \quad initial student model $q_{\theta_0}$ 
\State \quad \quad \ divergence $\mathcal{D}$ for training 
\State \quad \quad \ divergence $\mathcal{D}_{\text{JSD}}$ for discrepancy 
\State \quad \quad \ decay rate $\lambda$, \quad \ learning rate $\eta$

\Ensure Trained student model $q_{\theta_I}$

\For{$x$ in $(X, Y)$}

\While{\quad $y_t \neq EOS$} 
\State Compute threshold $\tau_t \gets \tau_0 e^{-\lambda t}$ 
\State Get student distribution $q_t \gets q_{\theta}(\cdot \mid y_{<t}, x)$ 
\State Get teacher distribution $p_t \gets p(\cdot \mid y_{<t}, x)$ 
\State Compute divergence $\mathcal{D}_{\text{JSD}} \gets \mathcal{D}_{\text{JSD}}(p_t \parallel q_t)$ 
\If{$\mathcal{D}_{\text{JSD}} \leq \tau_t$} 
\State Sample $y_t \sim q_t$ 
\Else 
\State Sample $y_t \sim p_t$ 
\EndIf 
% \State Update context $y_{<t+1} \gets y_{<t} \cup \{ y_t \}$ 

\EndWhile 
\State Update student model $\theta \gets \theta - \eta \nabla_{\theta} \mathcal{D}(p \parallel q_{\theta})(y|x)$ 
\EndFor 

\end{algorithmic} 
\end{algorithm}

%% file: text/4_experiments.tex
\subsection{Experimental Setup}
\paragraph{Implementation Details} 
For training, we randomly split \texttt{databricks-dolly-15k}~\cite{DatabricksBlog2023DollyV2} data into 14K samples for training and 500 samples for validation, following the experimental setup of \citet{minillm,ko2024distillm}. Following \citet{minillm} and \citet{ko2024distillm}, we incorporate a language modeling loss \cite{radford2018improving} using the OpenWebText ~\cite{Gokaslan2019OpenWeb} corpus in all experiments. Similar to \citet{gkd}, we use a mix of generated and ground-truth sequences, setting it at 0.5, to reduce the computational load during training. We use a decay factor of 0.1 for the main experiments. Further details can be found in Appendix~\ref{sec:appendix_training_details}.

\input{src/tab_main}

We evaluate our models on instruction-following tasks~\cite{ouyang2022training} across five datasets. We use Dolly~\cite{DatabricksBlog2023DollyV2}, comprising a 500-sample test set derived from \texttt{databricks-dolly-15k}. SelfInst~\cite{wang2022self}, contains 252 samples from a user-oriented instruction-following collection. Vicuna~\cite{chiang2023vicuna} provides 80 questions from its evaluation, spanning nine categories such as writing, roleplay, math, coding, and knowledge. S-NI~\cite{wang2022super} features a 9K test set from Super-Natural Instructions, with samples from a wide range of tasks. UnNI~\cite{honovich2022unnatural} consists of a 10K test set, randomly sampled from the core set of Unnatural Instructions. Similar to \citet{minillm}, we split the Dolly and SelfInst datasets into two subsets based on the ground truth response length for further analysis, as discussed in Section~\ref{sec:analysis_mitigating}
.

As the metric for evaluation, we report Rouge-L \cite{lin2004rouge} for all experiments. Following \citet{minillm} and \citet{ko2024distillm}, we also report GPT-4 feedback scores \cite{zheng2023judging} to DollyTest, SelfInst and VicunaEval for the results in Table ~\ref{tab:table_1}. We report the ratio of the total score of model responses and ground truth answers, using temperature 0.7 with evaluation prompt following \citet{zheng2023judging} and \citet{ko2024distillm}. 
Following \citet{ko2024distillm}, for all test sets we report the average score by sampling 5 responses with different seeds using temperature 1.0.

\paragraph{Models} 

To evaluate the effectiveness of SWITCH, we examine the instruction-following capabilities across three different pre-trained model families, which include GPT-2 \cite{radford2019language}, OPT \cite{zhang2022opt}, and OpenLLaMA-2 \cite{openlm2023openllama}.
For the GPT-2 model family, we use GPT-2 XL (1.5 billion parameters) as the teacher model, while the student models include GPT-2 Base (120M params), GPT-2 Medium (340M params), and GPT-2 Large (760M params).
Regarding the OPT model family, the teacher model is OPT-2.7B, and the student models are OPT-125M, OPT-350M, and OPT-1.3B.
In the OpenLLaMA-2 model family, the teacher model employs OpenLLaMA-2 with 7 billion parameters, while the student model uses OpenLLaMA-2 with 3 billion parameters.

Consistent with prior studies, we use the teacher model that is fine-tuned on the Dolly training set prior to knowledge distillation. The student model is similarly fine-tuned on the same dataset for three epochs. Evaluation is conducted using the models with the highest ROUGE-L score on the validation set. For the training of OpenLLaMA-2, we employ low-rank adaptation \cite{hu2021lora}, as in \citet{ko2024distillm}.

\paragraph{Baselines}
We conduct a comprehensive comparison of our method against a diverse set of baseline approaches, such as SFT, which directly fine-tunes the student model on the training data without involving knowledge distillation. KD \cite{hinton2015distilling} employs KLD using the teacher's output distributions as supervision at each token step. SeqKD \cite{kim2016sequence} applies SFT to sequences generated by the teacher model. GKD  \cite{gkd} utilizes Generalized JSD and leverages both ground-truth sequences and sequences generated by the student model. MiniLLM \cite{minillm} employs reverse KLD, using sequences generated by the student model and updating through policy gradient. DistiLLM \cite{ko2024distillm} uses skew reverse KLD and adaptively selects whether to use ground-truth sequences or those generated by the student model.

Compared to other methods, the distinctive aspect of SWITCH is its use of sequences generated not only by the student model but also with selective involvement from the teacher model. We employ reverse skew divergence with a weight of 0.1 for our main results, following the approach used in~\cite{ko2024distillm}, while also demonstrating the performance enhancement of SWITCH using different divergence losses in section~\ref{sec:analysis_loss}.

\subsection{Experimental Results}

\paragraph{Main Results}

Table~\ref{tab:table_1} illustrates the instruction-following performance, showing that SWITCH consistently outperforms state-of-the-art methods across various teacher-student configurations and evaluation metrics.
Notably, the performance gains are more pronounced in smaller student models that have a larger performance gap than the teacher model. This indicates that our method effectively bridges the gap between student and teacher models, especially when the student's capacity is significantly lower.
By outperforming various methods that utilize SGOs, we emphasize the importance of strategic teacher intervention during the distillation process, rather than relying on outputs generated solely by students. 
Full table results for OPT and OpenLLaMA-2 families can be found in the Appendix~\ref{sec:appendix_full_table}.

\paragraph{Performance on Different Response Lengths}

To further evaluate the effectiveness of our method, we analyze model performance across responses with varying ground truth lengths. In Figure~\ref{figure1}, we compare different distillation methods using SGOs against standard KD, which does not use SGOs. The experiment, conducted with GPT-2 Base (120M) and GPT-2 XL (1.5B), reports Rouge-L scores.

It is evident that methods that utilize SGOs, such as \citet{minillm} and \citet{ko2024distillm}, which are based on outputs generated exclusively by student models, experience a significant performance drop as the response length increases. In contrast, our SWITCH method maintains higher performance levels for instructions that require both short and long responses. This reduced performance degradation highlights the effectiveness of our approach in mitigating teacher misguidance in autoregressive sequence generation.

%% file: src/tab_main.tex
\begin{table*}[ht!]
\centering
\resizebox{\textwidth}{!}{%
\begin{tabular}{cclclclclcc}
\hline
\multirow{2}{*}{Model} &
  \multirow{2}{*}{Parameters} &
  \multirow{2}{*}{Method} &
  \multicolumn{2}{c}{Dolly} &
  \multicolumn{2}{c}{SelfInst} &
  \multicolumn{2}{c}{Vicuna} &
  S-NI &
  Unnatural \\ \cline{4-11} 
                            &                       &               & GPT4 & R-L & GPT4 & R-L & GPT4 & R-L & R-L  & R-L  \\ \hline
\multirow{20}{*}{GPT-2}                       & 1.5B                  & Teacher (SFT) & 49.2&     27.0 & 38.5&     15.5& 39.1&     16.2 & 25.2 & 33.4\\ \cline{2-11} 
                            & \multirow{7}{*}{120M}                  & SFT           & 34.1&     22.8 & 23.2&     10.3 & 24.4&     15.1 & 17.2 & 21.3 \\
                            &                       & KD            & 34.9&     23.2 & 23.7&     11.2 & 24.3&     15.9 & 17.7 & 22.2 \\
                            &                       & SeqKD         & 34.7&     23.7 & 23.8&     10.1 & 25.0&     15.0 & 15.0 & 23.2 \\
                            &                       & GKD           & 35.1&     23.7 & 25.1&     12.4 & 27.1&     17.2 & 23.1 & 26.5 \\
                            &                       & MiniLLM       & 35.2&     23.8 & 25.4&     12.7 & 27.2&     17.2 & 24.2 & 27.1 \\
                            &                       & DistiLLM      & 36.4&     25.6 & 26.1&     13.1 & 29.1&     18.4 & 27.5 & 28.0 \\
                            &                       & Ours          & \textbf{36.9}&     \textbf{26.2} & \textbf{26.9}&     \textbf{13.7}& \textbf{29.6}&     \textbf{18.6}& \textbf{28.1}& \textbf{28.4}\\ \cline{2-11} 
                            & \multirow{7}{*}{340M}                  & SFT           & 38.2&     24.1 & 27.5&     12.6 & 31.7&     15.6 & 24.2 & 27.3 \\
                            &                       & KD            & 38.1&     24.7 & 27.7&     12.9 & 31.9&     16.2 & 24.2 & 27.4 \\
                            &                       & SeqKD         & 38.6&     24.5 & 27.6&     13.1 & 32.1&     16.4 & 25.2 & 27.5 \\
                            &                       & GKD           & 40.1&     24.5 & 28.3&     14.4 & 33.3&     17.2 & 26.3 & 28.2 \\
                            &                       & MiniLLM       & 40.5&     25.1 & 28.1&     14.7 & 33.4&     17.5 & 26.8 & 29.4 \\
                            &                       & DistiLLM      & 41.4&     27.0 & 28.7&     15.1 & 34.1&     \textbf{18.2} & 28.2 & 30.1 \\
                            &                       & Ours          & \textbf{41.9}&     \textbf{27.4} & \textbf{29.0}&     \textbf{15.2} & \textbf{34.3}&     18.0& \textbf{28.5}& \textbf{30.2}\\ \cline{2-11} 
                            & \multirow{7}{*}{760M} & SFT           & 44.3&     25.4 & 33.1&     13.7 & 34.0&     16.5 & 23.2 & 27.5 \\
                            &                       & KD            & 44.6&     25.7 & 33.4&     14.1 & 34.5&     16.7 & 24.5 & 29.4 \\
                            &                       & SeqKD         & 44.6&     26.2 & 33.3&     14.6 & 34.5&     17.0 & 24.7 & 28.1 \\
                            &                       & GKD           & 46.1&     25.1 & 35.1&     14.9 & 35.1&     17.3 & 26.6 & 31.2 \\
                            &                       & MiniLLM       & 46.5&     26.4 & 35.2&     15.2 & 35.8&     17.9 & 28.0 & 32.3 \\
                            &                       & DistiLLM      & 47.1&     28.4 & 35.8&     15.5 & \textbf{36.2}&     \textbf{18.5} & 29.2 & 34.1 \\
                            &                       & Ours          & \textbf{47.3}&     \textbf{28.5} & \textbf{36.1}&     \textbf{15.7} & 36.1&     18.4 & \textbf{29.3} & \textbf{34.5} \\ \hline
\multirow{10}{*}{OPT}       & 2.7B                  & Teacher (SFT) & 48.7&     26.2 & 32.1&     13.3& 36.9&     16.6 & 23.4& 30.4\\ \cline{2-11} 
                            & \multirow{3}{*}{125M} & MiniLLM       & 31.2&     22.7 & 23.5&     10.1 & 24.1&     15.3 & 16.5 & 20.3 \\
                            &                       & DistiLLM      & 31.6&     24.9 & 24.4&     10.7 & 24.7&     16.1 & 21.4 & 23.2 \\
                            &                       & Ours          & \textbf{32.0}&     \textbf{25.1} & \textbf{24.9}&     \textbf{11.2} & \textbf{25.6}&     \textbf{16.5} & \textbf{23.1} & \textbf{24.5} \\ \cline{2-11} 
                            & \multirow{3}{*}{350M} & MiniLLM       & 36.1&     24.8 & 26.5&     13.2 & 28.2&     15.9 & 20.4 & 24.0 \\
                            &                       & DistiLLM      & 36.9&     25.1 & \textbf{27.1}&     \textbf{14.2} & 29.5&     16.8 & 22.0 & 25.7 \\
                            &                       & Ours          & \textbf{37.0}&     \textbf{25.4} & 26.9&     13.7 & \textbf{29.7}&     \textbf{17.0} & \textbf{22.3} & \textbf{26.1} \\ \cline{2-11} 
                            & \multirow{3}{*}{1.3B} & MiniLLM       & 43.1&     25.9 & 28.1&     14.3& 27.1&     16.6& 21.7 & 27.3 \\
                            &                       & DistiLLM      & 43.9&     26.8 & 29.4&     15.1& 28.1&     16.4& 24.5 & \textbf{30.4} \\
                            &                       & Ours          & \textbf{44.1}&     \textbf{27.0} & \textbf{29.5}&     \textbf{15.5}& \textbf{28.5}&     \textbf{16.9} & \textbf{24.9} & 30.2\\ \hline
\multirow{4}{*}{OpenLLaMA2} & 7B                    & Teacher (SFT) & 63.2&     28.8& 60.9&     20.5& 53.1&     17.1 & 34.8& 34.5\\ \cline{2-11} 
                            & \multirow{3}{*}{3B}   & MiniLLM       & 58.2&     27.4 & 57.1&     19.8 & 50.9&     19.4 & 35.4 & 36.2 \\
                            &                       & DistiLLM      & 58.9&     28.3 & 58.4&     19.7 & 52.0&     19.5 & 36.1 & 35.8 \\
                            &                       & Ours          & \textbf{59.4}&     \textbf{28.6}& \textbf{58.5}&     \textbf{20.0}& \textbf{52.1}&     \textbf{19.6}& \textbf{36.5}& \textbf{36.3}\\ \hline
\end{tabular}%
}
\caption{Evaluation results on 5 instruction-following datasets. Each GPT4 and ROUGE-L score is averaged over 5 random seeds. The best score for each model size is highlighted in \textbf{boldface}.}
\label{tab:table_1}
\vspace{-4mm}
\end{table*}

%% file: text/5_analysis.tex
\paragraph{Mitigating Misguidance in Lengthy SGOs}
\label{sec:analysis_mitigating}

In this section, we aim to determine whether teacher misguidance tends to increase with longer outputs and demonstrate that SWITCH can effectively mitigate this issue. To measure the extent of misguidance when using SGOs as training sequences, we calculate the correlation between Rouge-L scores of SGOs and training loss. To evaluate the impact of SWITCH on mitigating misguidance, we repeat the process, this time using the training loss obtained from sequences generated with SWITCH. We interpret students who require further tuning for the sample instruction based on the Rouge-L scores of the SGO—a lower coefficient indicates that students with an incorrect understanding of the given sample receive higher losses, while those with a correct understanding receive lower losses. 

In this experiment, we use GPT-2 Base (120M) at the start of the distillation process and GPT-2 XL (1.5B) as the teacher. To sample, we utilize the full Dolly train split for instructions. We use Spearman correlation for the scale-invariant measures \cite{zar2005spearman}.

\input{src/tab_cor}

The results in Table~\ref{tab:tab_cor} show that SGOs with longer sequences exhibit a greater tendency for misguidance. More importantly, they demonstrate that the SWITCH method effectively mitigates this misguidance, especially in longer SGOs. This underscores the crucial role of SWITCH in enhancing the training process by reducing teacher misguidance from lengthy SGOs.

\paragraph{Analysis of Teacher Intervention Strategies}
\label{sec:analysis_abl}
\input{src/tab_abl}

To analyze the key factors in mitigating misguidance, we conduct an ablation study to assess the impact of different teacher intervention strategies on the student model's performance. The following strategies are compared:

% \textit{Exponential Decaying Threshold (Ours)}: Uses an exponentially decaying divergence threshold for teacher intervention. 
\textit{Linear decrease}: To explore the necessity of exponential intervention scaling, we implemented a linearly decreasing threshold. This threshold diminishes linearly from a value of 1 to 0 across the maximum response length, offering a straightforward comparison to assess whether a more constant reduction in guidance affects performance differently.
\textit{Exponential growth}: Contrasting with our primary method, this strategy employs an exponentially increasing threshold, prioritizing early sequence token selection. This approach tests whether increased early incorporation in sequence generation helps the knowledge transfer.
\textit{Constant threshold}:  A threshold of 0.2 was used to determine if a fixed threshold could maintain or improve model performance compared to dynamic strategies. This constant threshold serves as a measure to examine the benefits of adaptive thresholds.
\textit{Teacher prob mix-in}:
To investigate an alternative to incorporate the teacher model during the generation, we compare it against a strategy that combines the sampling distributions of teacher and student models. This method, devised in \citet{minillm} to address reward hacking, employs a confidence-based probability weighting as in Equation~\eqref{eq9}. We define the mix-in strength $\alpha$ as 0.2, consistent with the settings from \citet{minillm}. This approach allows us to assess whether combining the probabilities can effectively train the student model compared to divergence-based thresholds.

% \begin{equation}
\begin{equation} \label{eq9}
\begin{split}
\Tilde{p}_t(y_t \mid y_{<t}, x) = \alpha \cdot p(y_t \mid y_{<t}, x) + \\ (1-\alpha) \cdot q_{\theta}(y_t \mid y_{<t}, x)
\end{split}
\end{equation}

\textit{Random Teacher Generation:} This strategy randomly selects between the teacher and student models for generating each token, exploring whether arbitrary teacher involvement might either benefit or hinder the knowledge transfer.

The results shown in Table ~\ref{tab:table_abl} indicate that our exponential decaying threshold strategy significantly outperforms other teacher intervention methods. This underscores the importance of dynamically adjusting teacher involvement based on the divergence between student and teacher distributions as the sequence progresses. Specifically, methods like the teacher probability mix-in that combines the teacher probabilities perform less effectively because they do not adjust teacher intervention based on divergence. By employing an exponentially decreasing threshold, our approach ensures that the teacher's guidance is applied when it is most beneficial, thereby enhancing knowledge transfer and effectively mitigating misguidance in the student model.

\paragraph{Performance Across Various Loss Functions}
\label{sec:analysis_loss}
To demonstrate the flexibility of our method, we experiment with various loss functions. Specifically, we compare performance both with and without applying our approach. We focus on various discrepancy measures commonly used in knowledge distillation for autoregressive models, including KLD, reverse KLD (RKL) \cite{hinton2015distilling,minillm}. We also experiment with Generalized JSD and skew reverse KLD (SRKL), using the weight 0.9 and 0.1 following \cite{gkd,ko2024distillm}.

\input{src/tab_loss}
As shown in Figure~\ref{figure_loss_temp}, the consistent improvements across various loss functions highlight the robustness and adaptability of our SWITCH method. This versatility allows SWITCH to handle a wide range of distillation objectives and remain compatible with emerging loss functions aimed at capturing complex teacher distributions. Its ability to generalize across tasks ensures reliable performance under varying distillation requirements, making it a valuable contribution to autoregressive model training.

\paragraph{Impact of Decay Factor on Performance and Token Generation Ratio}
\label{sec:analysis_decay}
To examine the impact of various decay factors in the exponential decay threshold on performance and the ratio of tokens generated by the student versus the teacher model, we conduct experiments using different decay factors. Table~\ref{table_decay} summarizes the results.

\input{src/tab_decay}

The decay factor of 1/10 yields the best performance, with a balanced proportion of tokens generated by the student and teacher models. This suggests that while our teacher intervention strategy offers significant advantages, excessive reliance on the teacher during training can negate the benefits of addressing the training-inference mismatch. Therefore, a cautious and balanced approach is necessary to optimize performance.

%% file: src/tab_cor.tex
\begin{table}[h]
\centering
\resizebox{\columnwidth}{!}{%
\begin{tabular}{lcc}
\hline
SGO token length & \multicolumn{2}{c}{Correlation coefficient ($\downarrow$)} \\ \hline
SWITCH           & \ding{55}          & \ding{51}           \\ \hline
1 $\sim$10         & -0.78       & -0.79       \\
11 $\sim$50         & -0.68       & -0.73       \\
51 $\sim$ inf.               & -0.59       & -0.69       \\ \hline
\end{tabular}%
}
\caption{Spearman's correlation coefficient between Rouge-L scores for SGOs and divergence loss, both with and without the use of SWITCH. Results are split by SGO token length. A lower coefficient suggests that incorrect students tend to have higher losses.}
\label{tab:tab_cor}
\end{table}

%% file: src/tab_abl.tex
\renewcommand{\arraystretch}{1.2}

\begin{table}[H]
\centering
\resizebox{0.90\columnwidth}{!}{%
\begin{tabular}{lccc}
\hline
\multirow{2}{*}{Intervention Strategy} & \multicolumn{3}{c}{Datasets}                                                      \\ \cline{2-4} 
                                 & Dolly & SelfInst & S-NI  \\ \hline
\textbf{Exponentially decaying threshold} & 26.2& 13.7& 28.1\\ \hline
w/  Linear decrease              & 25.1& 12.7& 27.1\\
w/ Exponential growth          & 23.4& 12.3& 24.8\\
Constant threshold               & 25.5& 12.9& 27.0\\
Teacher prob mix-in              & 24.6& 12.7& 26.3\\
Random teacher gen               & 24.8& 12.5& 25.2\\ \hline
\end{tabular}%
}
\caption{Comparison of the performance using different teacher involvement strategies.}
\label{tab:table_abl}
%\vspace{-10mm}
\end{table}

%% file: src/tab_loss.tex
\begin{figure}[ht]
\centering
\includegraphics[width=\columnwidth]{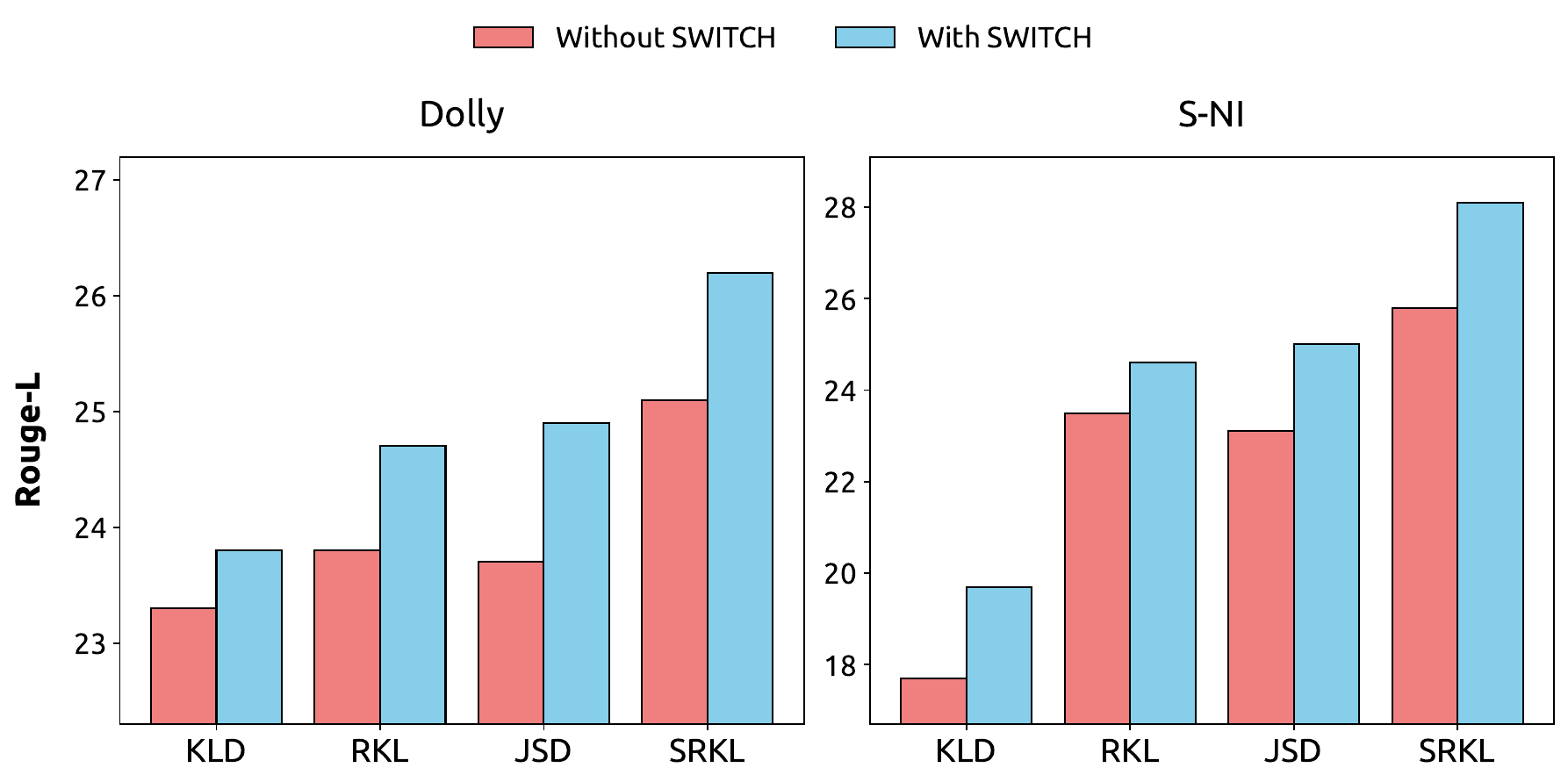} 
\caption{Application of our method SWITCH to different distillation loss}
\label{figure_loss_temp}

\end{figure} 

%% file: src/tab_decay.tex
\begin{table}[H]
\centering
\resizebox{\columnwidth}{!}{%
\begin{tabular}{lcllcc}
\hline
\multirow{2}{*}{Decay factor} & \multicolumn{3}{c}{Rouge-L} & \multicolumn{2}{c}{Tokens generated}        \\ \cline{2-6} 
     & Dolly & SelfInst & S-NI  & Student & Teacher \\ \hline
1/5  & 24.9& 12.4& 26.0& 22\%& 78\%\\
\textbf{1/10} & 26.2& 13.7& 28.1& 47\%& 53\%\\
1/15 & 26.0& 13.6& 27.6& 81\%& 19\%\\
1/25 & 25.7& 13.1& 27.5& 93\%& 7\%\\ \hline
                              & \multicolumn{1}{l}{}  &  &  & \multicolumn{1}{l}{} & \multicolumn{1}{l}{}
\end{tabular}%
}
\caption{Performance and token generation ratio using different decaying factor.}
\label{table_decay}
\end{table}

%% file: text/2_relatedworks.tex
\paragraph{KD for non-autoregressive models}
Knowledge Distillation (KD, \citealp{hinton2015distilling}) trains a smaller student model under the guidance of a more powerful teacher model, to have the student model replicate the teacher model's rich and complex representations. Various KD methods have been explored, including logit-based \cite{sun2019patient,sanh2019distilbert}, feature-based \cite{heo2019knowledge,chen2021cross} and relation-based \cite{park2019relational, wang2020minilm}. These methods are primarily applied to non-autoregressive architectures such as BERT and its variants for classification tasks.

\paragraph{KD for autoregressive models}

KD for autoregressive models can be broadly classified into black-box and white-box methods. Black-box methods resemble supervised fine-tuning (SFT), where the teacher model's generated outputs are used to train the student model \cite{hsieh2023distilling,jiang2023lion}. In contrast, white-box methods utilize the teacher model's output distribution, minimizing the discrepancy between the student and teacher distributions at each time step to provide training signals \cite{kim2016sequence}.

An area of development in white-box KD involves exploring various loss functions to better align the probability distributions between teacher and student models. Due to the complexity of generative tasks and the asymmetric nature of KLD, the student distribution may fail to capture the teacher distribution \cite{wen2023f}. To address these issues, alternative loss functions such as reverse-KLD and skew divergence have been employed \cite{minillm,ko2024distillm}. In this work, we conduct experiments to assess the compatibility of our method with these loss functions, a consideration given their impact on the effectiveness of the distillation process.

Another approach involves using student-generated outputs (SGOs) as the training sequence in the distillation process. This method aims to address training-inference mismatch—often referred to as exposure bias—where autoregressive models suffer from distribution mismatch between output sequences seen during training and those generated by the student during inference \cite{gkd}. By simulating the student's own learning trajectory, SGOs have been applied to various distillation methods \cite{gkd,minillm}. 
However, the use of SGOs introduces potential risks of misguidance from the teacher, due to the distribution mismatch between the student and teacher models. \citet{ko2024distillm} addresses this issue with an adaptive SGO scheduler, which mixes SFT with ground-truth data while gradually increasing the usage of SGOs. Our approach differs by directly adjusting SGOs based on discrepancies with the teacher model, addressing the accumulation of errors, which hinders effective knowledge transfer in long sequences \cite{arora2022exposure}. This work strategically incorporates the teacher into the generation process, ensuring the sequence is reliable from the teacher's perspective for accurate guidance while preserving the benefits of SGOs.

%% file: text/6_conclusion.tex
We propose SWITCH, a novel knowledge distillation approach that strategically incorporates the teacher model during the student's training sequence generation. By detecting discrepancies in token probabilities and applying an exponentially decaying threshold for selective teacher intervention, SWITCH balances the benefits of student learning from its own outputs while effectively mitigating teacher misguidance in long sequences. Comprehensive experiments across various model families and datasets show that SWITCH outperforms baseline methods, particularly in enhancing knowledge transfer for long sequences.

%% file: text/7_appendix.tex
\section{Additional Implementation Information}
\label{sec:appendix_training_details}

Our experimental setup largely follows the implementation of \cite{minillm, ko2024distillm}, with the exception of batch size. For models with fewer than 1B parameters, we search for learning rates in \{5e-4, 1e-4, 5e-5\} and batch sizes of \{8,16\}, training for 20 epochs. For models with more than 1B parameters, we search for learning rates in \{5e-5, 1e-5, 5e-6\} with a batch size of 8, training for 10 epochs. We exclude samples from \texttt{databricks-dolly-15k} that exceed the models' context length. For the generation of sequences during training, we use a temperature of 1.0.

To train OpenLLaMA2, we apply low-rank adaptation (LoRA) \cite{hu2021lora} to the query and value weights with a rank of 16, as described in \cite{ko2024distillm}. We conduct our experiments using 2 A100 80GB GPUs. Distilling OPT-2.7B from OPT-1.3B takes less than 12 hours.

For instruction-following prompts, we use the prompt shown in Figure~\ref{fig:sample_prompt} for both training and evaluation. For GPT-4 feedback scores, we use the prompt template in Figure~\ref{fig:gpt4prompt}, as outlined in \cite{zheng2023judging, ko2024distillm}. 

\begin{figure}[H]
    \centering
    \small
    \begin{tcolorbox}
    [width=\linewidth, sharp corners=all, colback=gray!10, boxrule=0.3mm]
    Below is an instruction that describes a task. \\
    Write a response that appropriately completes the request. \\

    \#\#\# Instruction: \\
    \{instruction\} \\

    \#\#\# Input: \\
    \{input\} \\

    \#\#\# Response:
    \end{tcolorbox}
    \vspace{-7pt}
    \caption{The prompt template for training and evaluation of instruction-following task experiments, following \citet{minillm}.}
    \label{fig:sample_prompt}
    \vspace{+150pt}
\end{figure}

\begin{figure}[H]
    \centering
    \small
    \begin{tcolorbox}
    [width=\linewidth, sharp corners=all, colback=gray!10, boxrule=0.3mm]
    [System] \\
    Please act as an impartial judge and evaluate the quality of the response provided by an AI assistant to the user question displayed below. Your evaluation should consider factors such as the helpfulness, relevance, accuracy, depth, creativity, and level of detail of the response. Begin your evaluation by providing a short explanation. Be as objective as possible. After providing your explanation, please rate the response on a scale of 1 to 10 by strictly following this format: ``[[rating]]'', for example: ``Rating: [[5]]''. \\

    [Question] \\
    \{question\} \\

    [The Start of Assistant’s Answer] \\
    \{answer\} 
    
    [The End of Assistant's Answer]
    \end{tcolorbox}
    \vspace{-7pt}
    \caption{The prompt template for evaluation from GPT-4 feedback following \citet{zheng2023judging} and \citet{ko2024distillm}.}
    \label{fig:gpt4prompt}
    \vspace{-10pt}
\end{figure}

\section{Training Time Analysis}
We provide a brief comparison of training times and Rouge-L scores across baselines utilizing SGOs in Table~\ref{tab:trainingtime}. This experiment is conducted with OPT-2.7B as the teacher model and OPT-1.3B as the student model for 10 epochs, using a batch size of 8 on 2 A100 GPUs. 

\input{src/tab_trainingtime}

\section{Qualitative Results} 
\label{sec:appendix_qualitative}

\input{src/tab_qual}
We present responses generated by models distilled using different methods based on GPT-2 Base (120M) models distilled from GPT-2 XL (1.5B) in Table~\ref{table_qualitative}. The prompts are sampled from \texttt{databricks-dolly-15k}. Our results show that SWITCH consistently produces more accurate responses compared to other knowledge distillation baselines.

\section{Full Table Results}
\label{sec:appendix_full_table}

The full results for the OPT family~\citet{zhang2022opt} are provided in Table~\ref{tab:main_opt}, and for the OpenLLaMA-2 family~\citet{openlm2023openllama} in Table~\ref{tab:main_llama}.

\input{src/tab_main_appendix_opt}

\input{src/tab_main_appendix_llama}

%% file: src/tab_trainingtime.tex
\begin{table}[H]
\centering
\resizebox{\columnwidth}{!}{%
\begin{tabular}{lccc}
\hline
\multicolumn{1}{c}{\multirow{2}{*}{Method}} & \multicolumn{1}{l}{\multirow{2}{*}{Relative Training Time}} & \multicolumn{2}{c}{Rouge-L against KD} \\ \cline{3-4} 
\multicolumn{1}{c}{} & \multicolumn{1}{l}{} & SelfInst & S-NI \\ \hline
GKD                  & 1x                   & 0.7      & 0.4  \\
MiniLLM              & 1.74x                & 0.9      & 0.9  \\
DistiLLM             & 0.64x                & 1.7      & 3.4  \\
SWITCH               & 1.84x                & 2.1      & 3.8  \\ \hline
\end{tabular}%
}
\caption{Relative training time and performance}
\label{tab:trainingtime}
\end{table}

%% file: src/tab_qual.tex
\begin{table*}[t]
\centering
\vspace{-10pt}

\vspace{5pt}
\renewcommand{\arraystretch}{1.4}
\resizebox{1.0\textwidth}{!}{
\addtolength{\tabcolsep}{2.5pt}
\begin{tabular}{m{3cm}m{20cm}}
\toprule[0.1em]
        \multicolumn{2}{c}{\textbf{\textit{Case \# 1}}} \\ \midrule[0.1em]
        
        Instruction & What is a dispersive prism? \\ \midrule
        
        Input & In optics, a dispersive prism is an optical prism that is used to disperse light, that is, to separate light into its spectral components (the colors of the rainbow). Different wavelengths (colors) of light will be deflected by the prism at different angles. $[$ 1 $]$ This is a result of the prism material's index of refraction varying with wavelength (dispersion). Generally, longer wavelengths (red) undergo a smaller deviation than shorter wavelengths (blue). The dispersion of white light into colors by a prism led Sir Isaac Newton to conclude that white light consisted of a mixture of different colors. \\ \hline
        Ground Truth & A dispersive prism is an optical prism that disperses the light's different wavelengths at different angles. When white light is shined through a dispersive prism it will separate into the different colors of the rainbow. \\ 

        \midrule \midrule
        SFT & A dispersive prism is a glass prism used toatters sunlight, that is, to reduce the blinding sunlight. \\
        \midrule
        KD  & A dispersive prism is a prism used to bring light into a single component. The dispersion of red light into colors by a lens produces a spectrum of colors similar to that of a prism and thus it can be described as a prism. \\
        \midrule
        SeqKD  & A dispersive prism is a optical prism that is used in the cinema to film light. \\
        \midrule
        GKD  & A dispersive prism is a light that occurs when the light from a red light source is scattered. It is also known as a rainbow \\
        \midrule
        MiniLLM  & A dispersive prism is used to provide effective infrared light scattering, which reduces incoming light as light travels out higher from tectonic plates in space. \\
        \midrule
        DistiLLM  & A dispersive prism is a prism that is used to distribute light, that is, against the background. \\
        \midrule
        \textbf{SWITCH}  & A dispersive prism is an optical prism that is used to distribute light, that is,to separate light into its spectral components (the colors of the Rainbow). \\ \midrule[0.1em]
        \multicolumn{2}{c}{\textbf{\textit{Case \# 2}}} \\ \midrule[0.1em]
        
        Instruction &  How long does it take for a mustard seed to germinate? \\ \midrule
        
        Input & Mustard seeds are the small round seeds of various mustard plants. The seeds are usually about 1 to 2 millimetres (0.039 to 0.079 in) in diameter and may be colored from yellowish white to black. They are an important spice in many regional foods and may come from one of three different plants: black mustard (Brassica nigra), brown mustard (B. juncea), or white mustard (Sinapis alba).Grinding and mixing the seeds with water, vinegar or other liquids creates the yellow condiment known as prepared mustard. Mustard seeds generally take eight to ten days to germinate if placed under the proper conditions, which include a cold atmosphere and relatively moist soil. \\ \hline

        Ground Truth & Mustard seeds generally take eight to ten days to germinate if placed under the proper conditions, which include a cold atmosphere and relatively moist soil \\
        
        \midrule \midrule
        SFT  & The yellowish-white to black color of a mustard seed usually starts about 1 to 2 millimeters (0.079 to 0.079) in diameter and may vary color to black or white depending on the plant. \\ 
        \midrule
        KD & It takes thousands of days to germinate a mustard seed. \\
        \midrule
        SeqKD  &  The length of mustard seeds depends on the type of mustard they are grown in.  A common type of mustard is yellow mustard, which is grown on the outer side of a mustard plant.  A yellow mustard seed can germinate into a white mustard.  A brown mustard seed can germinately germinate into a red mustard.\\
        \midrule
        GKD  & It takes about two to three days for a mustard seed to grow.  The seeds germinate from 1 to 2 millimetrons diameter each.  They can be colored from yellowish to black.  They are a key spice in many regional foods. \\
        \midrule
        MiniLLM  & The term mustard seed germinates about 1 to 2 months \\
        \midrule
        DistiLLM  & The time between germination and the onset of yellowish-white-to-black is about 1-2 months. \\
        \midrule 
        \textbf{SWITCH}  &  Mustard seeds take about 8 to 10 days to germinate. \\ 

\bottomrule[0.1em]

\end{tabular}
}
\caption{Instruction-following cases from the \texttt{databricks-dolly-15k}~\cite{DatabricksBlog2023DollyV2} dataset.}
\label{table_qualitative}
\end{table*}

%% file: src/tab_main_appendix_opt.tex
\begin{table*}[ht!]
\centering
\resizebox{\textwidth}{!}{%
\begin{tabular}{cclclclclcc}
\hline
\multirow{2}{*}{Model} &
  \multirow{2}{*}{Parameters} &
  \multirow{2}{*}{Method} &
  \multicolumn{2}{c}{Dolly} &
  \multicolumn{2}{c}{SelfInst} &
  \multicolumn{2}{c}{Vicuna} &
  S-NI &
  Unnatural \\ \cline{4-11} 
                            &                       &               & GPT4 & R-L & GPT4 & R-L & GPT4 & R-L & R-L  & R-L  \\ \hline
\multirow{20}{*}{OPT}& 2.7B& Teacher (SFT) & 48.7&     26.2 & 32.1&     13.3& 36.9&     16.6 & 23.4& 30.4\\ \cline{2-11} 
                            & \multirow{7}{*}{125M}& SFT           & 29.1&     20.5& 21.9&     8.7& 19.7&     13.5& 15.7& 17.4\\
                            &                       & KD            & 28.7&     22.1& 22.3&     8.9& 20.1&     14.7& 16.5& 18.4\\
                            &                       & SeqKD         & 29.4&     21.4& 22.4&     8.4& 19.6&     14.6& 17.4& 18.1\\
                            &                       & GKD           & 30.1&     22.5& 23.1&     9.5& 22.3&     15.1& 19.4& 21.4\\
                            &                       & MiniLLM       & 31.2&     22.7 & 23.5&     10.1 & 24.1&     15.3 & 16.5 & 20.3 \\
                            &                       & DistiLLM      & 31.6&     24.9 & 24.4&     10.7 & 24.7&     16.1 & 21.4 & 23.2 \\
                            &                       & Ours          & \textbf{32.0}&     \textbf{25.1} & \textbf{24.9}&     \textbf{11.2} & \textbf{25.6}&     \textbf{16.5} & \textbf{23.1} & \textbf{24.5} \\ \cline{2-11} 
                            & \multirow{7}{*}{350M}& SFT           & 33.7&     23.1& 24.3&     11.9& 25.1&     15.1& 18.8& 21.6\\
                            &                       & KD            & 33.6&     23.5& 24.8&     12.3& 27.0&     16.1& 18.9& 21.5\\
                            &                       & SeqKD         & 35.3&     23.8& 24.1&     12.8& 27.2&     15.3& 19.5& 22.4\\
                            &                       & GKD           & 35.9&     23.9& 26.3&     13.6& 27.3&     16.3& 19.6& 24.4\\
                            &                       & MiniLLM       & 36.1&     24.8 & 26.5&     13.2 & 28.2&     15.9 & 20.4 & 24.0 \\
                            &                       & DistiLLM      & 36.9&     25.1 & \textbf{27.1}&     \textbf{14.2} & 29.5&     16.8 & 22.0 & 25.7 \\
                            &                       & Ours          & \textbf{37.0}&     \textbf{25.4} & 26.9&     13.7 & \textbf{29.7}&     \textbf{17.0} & \textbf{22.3} & \textbf{26.1}\\ \cline{2-11} 
                            & \multirow{7}{*}{1.3B}& SFT           & 40.1&     24.8& 25.1&     13.1& 26.4&     15.5& 20.8& 27.0\\
                            &                       & KD            & 40.7&     25.1& 26.1&     13.4& 26.8&     15.4& 21.1& 27.2\\
                            &                       & SeqKD         & 41.5&     26.1& 26.2&     12.8& 26.1&     15.7& 21.1& 26.6\\
                            &                       & GKD           & 43.1&     25.8& 27.6&     14.1& 27.2&     16.1& 21.5& 27.4\\
                            &                       & MiniLLM       & 43.1&     25.9 & 28.1&     14.3& 27.1&     16.6& 21.7 & 27.3 \\
                            &                       & DistiLLM      & 43.9&     26.8 & 29.4&     15.1& 28.1&     16.4& 24.5 & \textbf{30.4} \\
                            &                       & Ours          & \textbf{44.1}&     \textbf{27.0} & \textbf{29.5}&     \textbf{15.5}& \textbf{28.5}&     \textbf{16.9} & \textbf{24.9} & 30.2\\ \hline

\end{tabular}%
}
\caption{Evaluation results on 5 instruction-following datasets with OPT model family~\cite{zhang2022opt}. Each GPT4 and ROUGE-L score is averaged over 5 random seeds. The best score for each model size is highlighted in \textbf{boldface}.}
\label{tab:main_opt}
\end{table*}

%% file: src/tab_main_appendix_llama.tex
\begin{table*}[ht!]
\centering
\resizebox{\textwidth}{!}{%
\begin{tabular}{cclclclclcc}
\hline
\multirow{2}{*}{Model} &
  \multirow{2}{*}{Parameters} &
  \multirow{2}{*}{Method} &
  \multicolumn{2}{c}{Dolly} &
  \multicolumn{2}{c}{SelfInst} &
  \multicolumn{2}{c}{Vicuna} &
  S-NI &
  Unnatural \\ \cline{4-11} 
                            &                       &               & GPT4 & R-L & GPT4 & R-L & GPT4 & R-L & R-L  & R-L  \\ \hline
\multirow{8}{*}{OpenLLaMA2}& 7B& Teacher (SFT) & 63.2&     28.8& 60.9&     20.5& 53.1&     17.1 & 34.8& 34.5\\ \cline{2-11} 
                            & \multirow{7}{*}{3B}& SFT           & 51.3&     25.3& 50.2&     16.5& 41.5&     15.7& 29.7& 30.1\\
                            &                       & KD            & 52.5&     26.1& 51.7&     16.6& 43.1&     16.1& 30.1& 32.1\\
                            &                       & SeqKD         & 53.1&     25.8& 53.1&     16.3& 46.1&     16.5& 30.5& 31.1\\
                            &                       & GKD           & 57.3&     27.0& 57.2&     19.2& 49.4&     19.5& 34.3& 35.4\\
                            &                       & MiniLLM       & 58.2&     27.4 & 57.1&     19.8 & 50.9&     19.4 & 35.4 & 36.2 \\
                            &                       & DistiLLM      & 58.9&     28.3 & 58.4&     19.7 & 52.0&     19.5 & 36.1 & 35.8 \\
                            &                       & Ours          & \textbf{59.4}&     \textbf{28.6}& \textbf{58.5}&     \textbf{20.0}& \textbf{52.1}&     \textbf{19.6}& \textbf{36.5}& \textbf{36.3}\\ 
                            \hline

\end{tabular}%
}
\caption{Evaluation results on 5 instruction-following datasets with OpenLLaMA2 family~\cite{openlm2023openllama}. Each GPT4 and ROUGE-L score is averaged over 5 random seeds. The best score for each model size is highlighted in \textbf{boldface}.}
\label{tab:main_llama}
\end{table*}